\pdfoutput=1

\documentclass[11pt]{article}

\usepackage{natbib}
\setcitestyle{numbers,square}
\usepackage{CUSTOM}

\usepackage{times}
\usepackage{latexsym}
\usepackage[T1]{fontenc}
\usepackage[utf8]{inputenc}
\usepackage{microtype}
\usepackage{inconsolata}
\usepackage{algorithm}
\usepackage{algorithmic}
\usepackage{amsmath}
\usepackage{graphicx}
\usepackage{multirow}
\usepackage{subcaption}
\usepackage{booktabs}

\graphicspath{ {./figure} }

\title{Automatic Extraction of Relationships among Motivations, Emotions and Actions from Natural Language Texts}

\author{Fei Yang \\
         CogBeauty Lab \\
         yftadyz@163.com}

\begin{document}
\maketitle
\begin{abstract}
We propose a new graph-based framework to reveal relationships among motivations, emotions and actions explicitly given natural language texts. A directed acyclic graph is designed to describe human's nature. Nurture beliefs are incorporated to connect outside events and the human's nature graph. No annotation resources are required due to the power of large language models. Amazon Fine Foods Reviews dataset is used as corpus and food-related motivations are focused. Totally 92,990 relationship graphs are generated, of which 63\% make logical sense. We make further analysis to investigate error types for optimization direction in future research.

\end{abstract}

\section{Introduction}

In daily life, different motivations drive humans to produce different behaviors, and at the same time, the satisfaction of motivations leads to different emotions. Understanding relationships among motivations, emotions, and subsequent actions has drawn a lot of attentions in the research community. One prevailing practice is, given an event text, annotators generate description texts of its motivations, emotions, and subsequent actions \citep{rashkin2018event2mind, sap2019atomic, ghosal2022cicero}. Or annotators mark motivations, emotions, and actions on its contexts \citep{poria2021recognizing, mostafazadeh2020glucose, gui2018event}. Then deep learning models are trained over the generated or labeled datasets, which encode the relationships into the models' parameters. One drawback of this paradigm is, it fails to reveal relationships explicitly, providing not much help in understanding human intelligence, although these black-box models perform well in real applications. Another drawback is, it heavily relies on human resources and workflow designs for annotation.

\begin{figure*}[h!]
  \includegraphics[width=2\columnwidth]{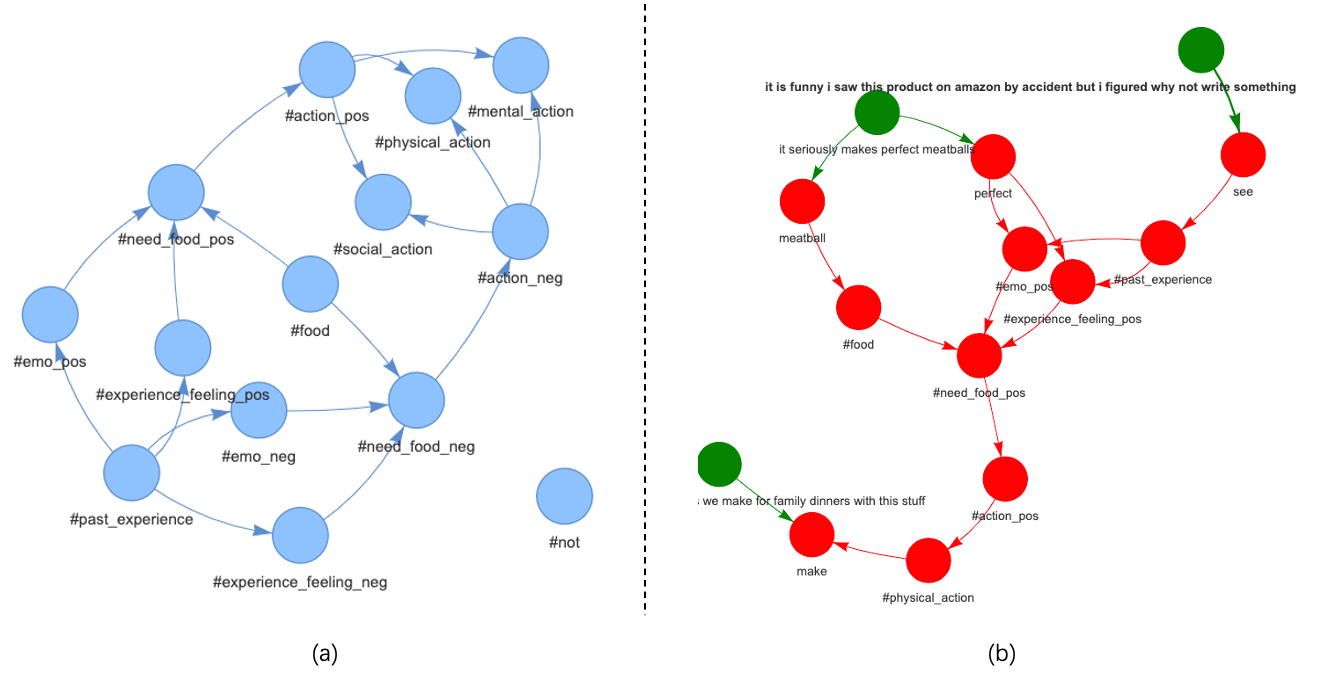}
  \caption{(a) Nature Design. This graph reveals the interactive mechanism among motivations, emotions and actions in human's nature. (b) An MEA-DAG example. Events are presented in green color and the activated nodes of Nature Design are in red color. Other nodes are omitted. Nurture Belief participates in linking events to corresponding nodes, and explicitly presented in the MEA-DAG. For instance, "it seriously makes perfect meatballs", has a connection with \texttt{\#food} by the belief tuple ("meatball", \texttt{\#food}).}
  \label{fig:demo}
\end{figure*}

In this work, we propose a framework to explicitly handle relationships among motivations, emotions and actions, which automatically generates directed acyclic graphs (MEA-DAG) given natural language texts. By drawing on findings from cognitive science, a Nature Design graph is built manually, which reveals human's inside nature, being formed through thousands of years of genetic evolution. Our framework also incorporates Nurture Belief, learned from developmental experiences. Nurture Belief plays a key role in connecting outside world events and Nature Design. Figure~\ref{fig:demo} shows the Nature Design graph and a MEA-DAG example. Large language model (LLM) is used to extract and improve the quality of Nurture Belief. Therefore no annotation resources are required in our framework, and efforts are put on prompt engineering instead of annotation workflow design. 

Review texts from Amazon Fine Foods Reviews dataset \citep{mcauley2013amateurs} are used as corpus. To reduce the complexity of the problem, only the motivation of human's need for food \citep{maslow1943theory} is focused. We divide this motivation into two types: positive and negative, which correspond to food need being met and not met respectively. Totally 92,990 reviews have valid MEA-DAGs and 63\% of them make logical sense. Error analysis is implemented to investigate the error types and find future research directions.

\section{Related Work}

Event2Mind \citep{rashkin2018event2mind} asks annotators to provide short textual descriptions of motivations and emotional reactions given an event. The collected texts serve as the training set of a encoder-decoder model, which predicts motivation and emotion over a new event. ATOMIC \citep{sap2019atomic} extends the annotation dimensions, and trains a encoder-decoder model for inference. In \citep{ghosal2022cicero}, given an event in a dialogue, annotators answer five dimensions: cause, subsequent event, prerequisite, motivation, and emotional reaction for tuning transformer-based models. Instead of generating artificial answers, other researchers choose to mark key information directly on the contexts of events. \citet{poria2021recognizing} annotates the cause of emotions manually at phrase level in two conversation datasets, and transformer models are fine-tuned for inference. \citet{gui2018event} annotates the cause of emotions from emotional context, and then a convolution kernel-based model is learned. In \citep{mostafazadeh2020glucose}, given a sentence and its context, ten dimensions including motivation, emotion, other implicit causes, and its effect are annotated by crowdsourcing. They then train a encoder-decoder model to infer both specific statements and general rules for new scenarios. All these methods code the relationship among motivations, emotions and actions into parameters of their models in a black-box style. Therefore, they contribute very little to the understanding of this relationship, although their models could do excellent inference on new scenarios.

\section{Methods}
\begin{figure*}[h!]
  \includegraphics[width=2\columnwidth]{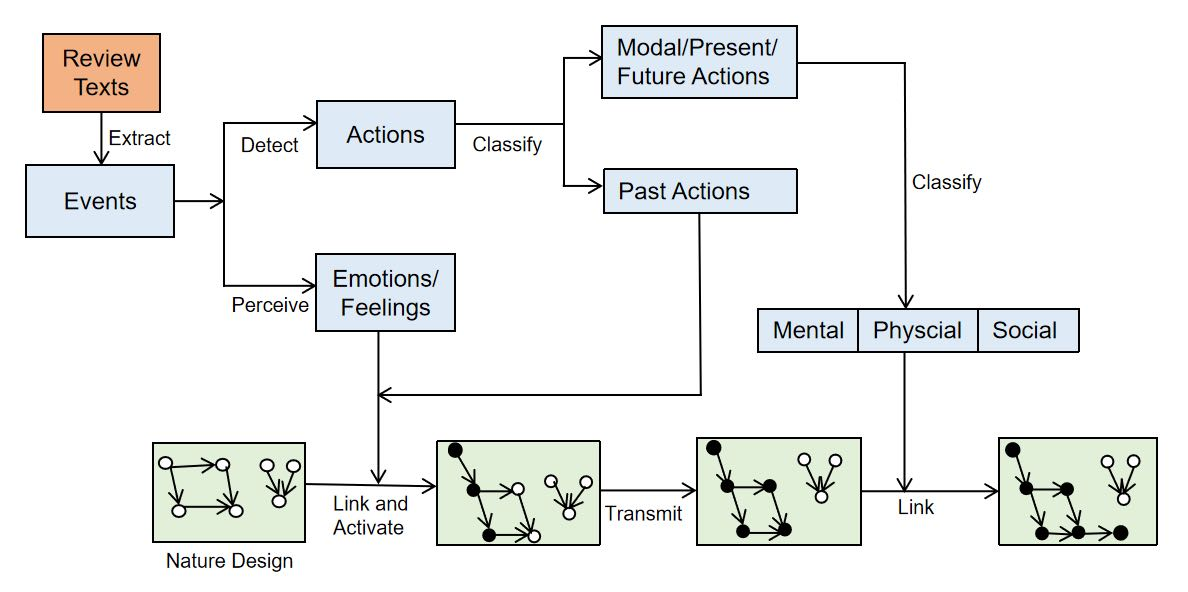}
  \caption{Framework of computing a MEA-DAG. It inputs review texts (top left corner) and outputs a graph (bottom right corner). The green line at the bottom shows the evolution of a MEA-DAG in different processing stages. }
  \label{fig:frame}
\end{figure*}

Our framework consists of four phases: (1) loading Nature Design and Nurture Belief, (2) perceiving states, (3) forward transmitting, and (4) taking actions, which are shown in Figure~\ref{fig:frame}. It mimics human brain's cognition process. First of all, a brain stores innate evolutional design and acquired developmental experiences. Next, suppose an event occurs around, the brain perceives this event, then neutrons send signals along axons and dendrites, and finally an action is taken by invoking body parts. Afterwards, the brain perceives feedback from the action, which forms a closed loop of cognition, providing abilities to (1) form new knowledge about this world, and (2) guide next action based on all existed knowledge in the brain. The feedback loop is not covered in this work and left for future research.

\subsection{ Loading Nature Design and Nurture Belief } 

Nature Design starts from \texttt{\#past\_experience}, whose behavioral outcomes drive emotions and feelings. Emotions and feelings are involuntary, which serve as passive states, reflecting patterns of physiological activities \citep{panksepp2004affective}. After perceiving these passive states, we infer whether human's need of food is satisfied or not. Positive feelings or emotions mean the need is satisfied while negatives mean the opposite, which are shown as directed links in Figure~\ref{fig:demo} (a). Positive actions are driven to strengthen being able to continuously meet the need when it's satisfied. Negative actions are driven to prevent it from happening again when human's need is dissatisfied. All actions are further broken down into three types: \texttt{Mental}, \texttt{Physical} and \texttt{Social}. Nodes of Nature Design are regarded as innate, different from learned experience \citep{izard1992basic,deci2000and}. Node definitions are summarized in Table~\ref{tab:nature_nodes}. We admit that actions are not only determined by motivations, but also biologically, culturally, and situationally determined as well \citep{maslow1958dynamic}. We ignore these factors and explore them in future research.

\begin{table}[t]
			\centering
			\footnotesize
			\caption{Explanation of nodes in Nature Design.}
			\label{tab:nature_nodes}
			{\
			\begin{tabular}{p{0.18\textwidth}|p{0.25\textwidth}}
				\toprule 
				Node & Explanation  \\
				\midrule
				\texttt{\#food} & Food entities, e.g. bread, apple. \\
				\midrule
				\texttt{\#experience\_feeling \newline \_pos} & Positive feelings, e.g. delicious, easy. \\
                                  \texttt{\#experience\_feeling \newline \_neg} & Negative feelings, e.g. bitter, hard. \\
                                  \midrule
                                \texttt{\#emo\_pos} &Positive emotions, e.g. happy, cheerful. \\
                                \texttt{\#emo\_neg} &Negative emotions, e.g. sad, angry. \\
                                \midrule
                                 \texttt{\#need\_food\_pos} &Human's need of food is satisfied. \\ 
         			\texttt{\#need\_food\_neg} &Human's need of food is dissatisfied. \\
				\midrule
         			\texttt{\#past\_experience}& Actions that take place in the past and result in a change of need state, e.g. bought, searched. It's the root node of Nature Design. \\
				\midrule
				\texttt{\#action\_pos} &Actions that are driven to strengthen being able to continuously meet the need. \\
         			\texttt{\#action\_neg} &Actions that are driven to prevent it from happening again when human's need is dissatisfied. \\
         			\midrule	
				\texttt{\#mental\_action} &Actions that happen inside human beings, not visible, e.g. analyze, versify. \\
        				\texttt{\#physical\_action} &Actions that happen outside human beings and are visible, e.g. wash, peel. \\
         			\texttt{\#social\_action} &Actions that are directed at others, e.g. denounce, rent. \\
				\bottomrule
			\end{tabular}
			}
        \end{table}

Nurture Belief includes three parts: food entities, experience feelings and emotions. They connect real world descriptions to abstractive concepts, stored as a set of tuples \{("word", \texttt{node})\}. For instance, "cheerful" describes a positive emotion, which is stored as ("cheerful", \texttt{\#emo\_pos}). When an event is linked to a node, the related Nurture Belief tuples are explicitly presented in its MEA-DAG.


\textbf{Food Entities. } WordNet \citep{miller1995wordnet} provides an ability to link concrete entities to abstract categories. We start from the word "food" and find all its hyponyms. The hyponyms with food-unrelated senses are removed to improve accuracy. Totally 1,842 tuples of ("word", \texttt{\#food}) are collected. 

\begin{figure}[h!]
  \includegraphics[width=1\columnwidth]{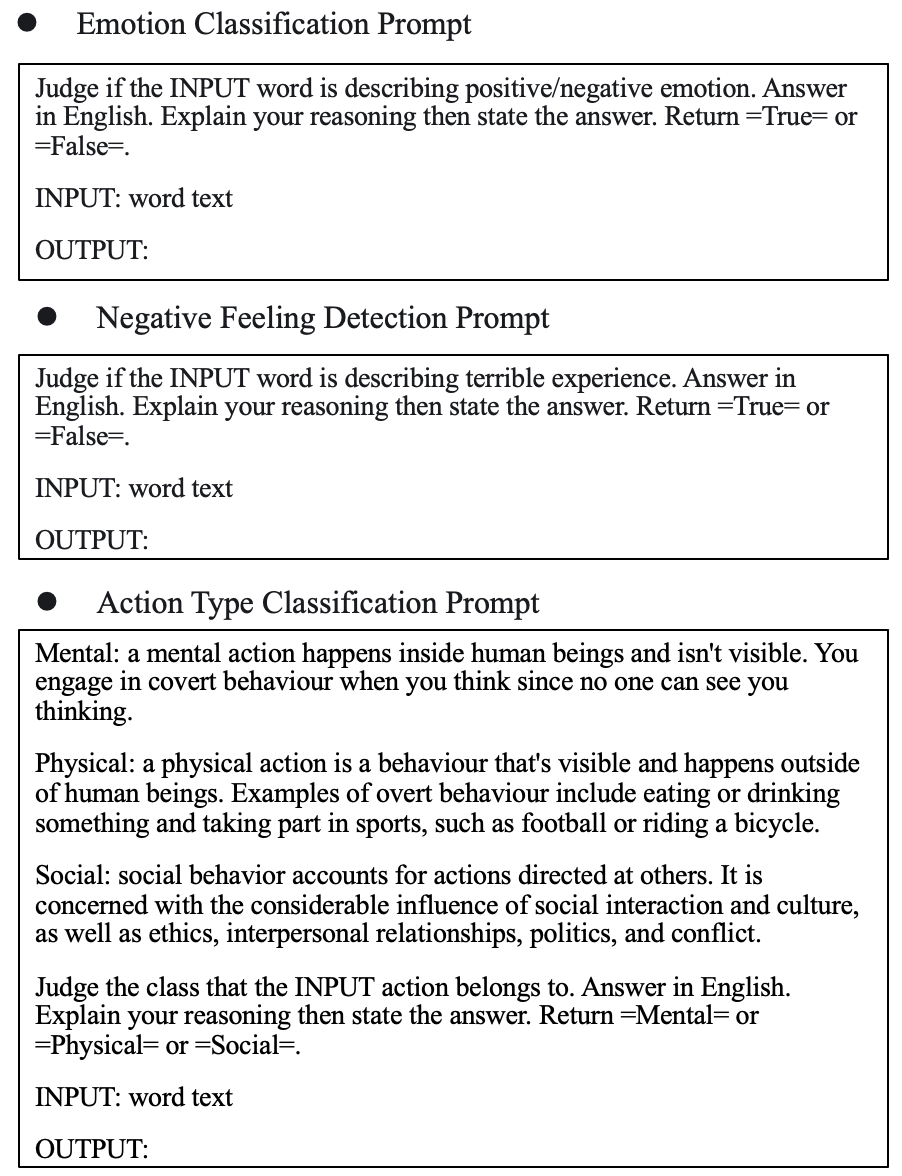}
  \caption{Prompt engineering. We rely on LLM to accelerate the establishment of Nurture Belief, and no longer rely on manual labeling resources.}
  \label{fig:prompt}
\end{figure}

\textbf{Experience Feelings. } SentiWordNet \citep{baccianella2010sentiwordnet} provides positive, negative and neutral feeling scores at sense level. By setting \texttt{PosScore}>0.6 and \texttt{NegScore}>0.6, positive and negative senses are extracted out respectively. Adjectives with only positive senses are classified as \texttt{\#experience\_feeling\_pos}, and negative-senses only are classified as \texttt{\#experience\_feeling\_neg}. GLM-4 \citep{glm2024chatglm}, an open-source LLM, is used over the negative adjectives to filter out bad cases. We list the prompt in Figure~\ref{fig:prompt}. Totally 1,415 positive tuples and 1,239 negative tuples are collected.


\textbf{Emotions. } \citet{shaver1987emotion} identify 135 base words which belong to six primary emotion classes: \texttt{Anger}, \texttt{Fear}, \texttt{Joy}, \texttt{Love}, \texttt{Sadness}, and \texttt{Surprise}. Synonyms of the base words are searched manually as extension words \footnote{Synonym Website: \url{https://www.merriam-webster.com/thesaurus}}. Extension words are classified as the same emotion class as their corresponding base words. We only keep adjectives and verbs. The base and extension words which belong to \texttt{Joy} and \texttt{Love} are classified as \texttt{\#emo\_pos}, and words from \texttt{Anger}, \texttt{Fear} and \texttt{Sad} are \texttt{\#emo\_neg}. GLM-4 is used to filter out bad cases, and its prompt is listed in Figure~\ref{fig:prompt}. Totally 1,425 positives tuples and 1,946 negatives tuples are collected. 


\subsection{Perceiving States } 
\label{perceiving}
ASER \citep{ZhangLPKOFS22, ZhangLPSL20} is used to extract events from reviews texts. By resorting to POS tagging \footnote{Penn Treebank POS tags. Check more details in \url{https://www.ling.upenn.edu/courses/Fall_2003/ling001/penn_treebank_pos.html}} and dependency parsing, we detect the following keyword combinations in an event: (1) food entity + feeling state, (2) food entity + emotion state, (3) "I/We" + emotion state, and (4) emotional action. These combinations indicate mental states about food. Keywords link and activate corresponding nodes in Nature Design, like "meatball" and "perfect" in Figure~\ref{fig:demo} (b). If "not" appears in an event, then feeling and emotion keywords would link and activate opposite nodes. For instance, in the event "I am not happy", "happy" is linked to \texttt{\#emo\_neg} rather than \texttt{\#emo\_pos}. 

\subsection{Forward Transmitting. }
Links in a MEA-DAG indicate the direction of signal transmission. The node pointed by a link is tail node, and the node on the other side is head node. Activated nodes in \ref{perceiving} send out signals along links to tail nodes, which is a forward transmitting process. For example, in Figure~\ref{fig:demo} (a), when the node \texttt{\#emo\_pos} is activated, as head node, it sends out a signal which activates its tail node \texttt{\#need\_food\_pos}, and then \texttt{\#action\_pos} is activated by its head node \texttt{\#need\_food\_pos}.

\subsection{Taking Actions. }
Action events are detected according to patterns listed in Table~\ref{tab:action_pattern}. Only events that have first-person subject "I/We" are considered. Next a \texttt{Past} event is determined by checking if the POS tagging of its verb is \texttt{VBD} or \texttt{VBN}, which is then linked to and activates the node \texttt{\#past\_experience}. Other events, \texttt{Modal/Present/Future}, are further classified into three types: \texttt{Mental}, \texttt{Physical} and \texttt{Social} by GLM-4. Definitions and prompts are listed in Figure~\ref{fig:prompt}. Events of each type are linked to the following activated nodes respectively, \texttt{\#mental\_action}, \texttt{\#physical\_action} or \texttt{\#social\_action}.

\begin{table}[t]
	\footnotesize
	\centering
	\caption{ Action event patterns. Only first-person subject patterns are considered. We refer to the ASER pattern writing format. 
	} \label{tab:action_pattern}	   
	{
		\begin{tabular}{p{0.2\textwidth}|cp{0.08\textwidth}}
			\toprule 
			Pattern & Example \\
			\midrule
			$I/We$\ -nsubj-$v_1$ & "I freeze" \\
			$I/We$\ -nsubj-$v_1$-dobj-$n_2$ & "I slice the loaf" \\
			$I/We$\ -nsubj-$v_1$-xcomp-$a$ & "I feel hungry" \\
			$I/We$\ -nsubj-($v_1$-iobj-$n_2$)-dobj-$n_3$ & "I give this product 5 star"\\

			$I/We$\ -nsubj-$v_1$-xcomp-$a_1$-cop-$be$ & "I expect to be served" \\
			$I/We$\ -nsubj-$v_1$-xcomp-$n_2$-cop-$be$ & "I want to be a gourmet" \\
			$I/We$\ -nsubj-$v_1$-xcomp-$v_2$-dobj-$n_2$ &"I wait to pay the product" \\
			$I/We$\ -nsubj-$v_1$-xcomp-$v_2$ &"I want to cook" \\

			$I/We$\ -nsubj-$v_1$-nmod-$n_2$-case-$p_1$ & "I go into the kitchen" \\
			($I/We$\ -nsubj-$v_1$-dobj-$n_2$)-nmod-$n_3$-case-$p_1$ & "I push the pizza into the oven" \\

			\bottomrule
		\end{tabular}
	}
\end{table}

\begin{table*}[tb]
\centering
\begin{tabular}{@{\hspace{5pt}}l@{\hspace{5pt}}l@{\hspace{5pt}}r@{\hspace{5pt}}r@{}c@{\hspace{5pt}}r@{\hspace{5pt}}r@{}c@{\hspace{5pt}}r@{\hspace{5pt}}r}
	\multicolumn{2}{c}{\multirow{2}{*}{}} & \multicolumn{2}{c}{Test} & & \multicolumn{2}{c}{Short-Test} & & \multicolumn{2}{c}{Long-Test} \\
	\cmidrule{3-4} \cmidrule{6-7} \cmidrule{9-10}
	 & & Count & Percent & & \quad Count & Percent \quad& & \quad Count & Percent \quad\\
	\midrule
	\multicolumn{1}{l|}{\multirow{2}{*}{Sample}}  &\  Incorrect & 37 & 37.0\% & & 14 & 29.8\%& &23&43.4\% \\
	 \multicolumn{1}{l|}{} & \ Total&100&100.0\%& &47&100.0\%& &53&100.0\% \\
	 \midrule
	 \midrule
	 \multicolumn{1}{l|}{\multirow{8}{*}{Error}} & \ Event Linking Loss &	9&	21.4\%&&	3	&20.0\%&&	6	&22.2\% \\	
	 \multicolumn{1}{l|}{}&\  ASER Extraction Loss	&7	&16.7\%&&	2&	13.3\%&&	5&	18.5\% \\
	 \multicolumn{1}{l|}{}&  \ Wrong Subsequent Action &	6	&14.3\%&&	1&	6.7\%	&&5&	18.5\% \\
	 \multicolumn{1}{l|}{}&\  Word Sense Ambiguity	&6	&14.3\%&&	3&	20.0\%&&	3&	11.1\% \\
	 \multicolumn{1}{l|}{}&\  Wrong Belief	&6&	14.3\%&&	4&	26.7\%&&	2&	7.4\%\\
	 \multicolumn{1}{l|}{}&\  Wrong Past Action&	5&	11.9\%&&	1&	6.7\%&&	4&	14.8\%\\
	 \multicolumn{1}{l|}{}&\  Negation Loss&	3&	7.1\%&&	1&	6.7\%&&	2&	7.4\%\\
	 \multicolumn{1}{l|}{}& \ Total&	42&	100.0\%&&	15&	100.0\%&&	27&	100.0\%\\
	\bottomrule
\end{tabular}
\caption{Accuracy and error types with count and percentage distribution. We present the results for the test set, the short-test set (sentence number < 5) and the long-test set (sentence number $\geq$ 5).}
\label{eval_error_anal}
\end{table*}

\section{Evaluation}
\subsection{Error Analysis}
Totally 92,990 valid MEA-DAGs are extracted out from 568,454 reviews. A valid MEA-DAG is defined as only \texttt{\#need\_food\_pos} or \texttt{\#need\_food\_neg} is activated. From valid MEA-DAGs, 100 samples are randomly chosen as a test set for manual evaluation of correctness. A MEA-DAG is incorrect if the revealed relationship is not logically making sense. During the evaluation, seven types of errors are found, whose distribution is shown in Table~\ref{eval_error_anal}. Totally 42 errors are detected and 37 samples having at least one error. Event Linking Loss and ASER Extraction Loss are the top two error types on the test set. In this section, we discuss why each error type occurs and possible solutions to improve it. Detailed MEA-DAG examples of each error type are appended in Appendix \ref{sec:appendix_a}.

\textbf{Event Linking Loss. } A MEA-DAG fails to incorporate critical information of events, making the revealed relationship not logically complete. The reasons lie in the coarseness of Nature Design, capturing very limited concepts. The limited number of patterns for linking events and nodes also leads to this loss. Besides adding more nodes and patterns, one interesting direction of improvement is to equip the algorithm with learning ability, being able to automatically build new nodes and links when it perceives new events and their outcomes.

\textbf{ASER Extraction Loss. } ASER fails in extracting out critical events from review texts, breaking the logic completeness. This happens due to the limited patterns of event-extractions by ASER. For instance, "I do not know how I could say whether or not the cat food is tasty" would be extracted as one event, "the cat food is tasty", missing key phrases "do not know" and "whether or not". It's necessary to find a method of understanding a sentence as a whole.

\textbf{Wrong Subsequent Action. } An action event should not be linked to the children of \texttt{\#action\_pos} or \texttt{\#action\_neg}, as it's not driven by human's need. For example, in the event "I consider myself a pro when it comes to popcorn", the reason of being a pro is not triggered by one specific satisfaction of food-related need, but the rich experience of eating popcorn. This kind of error happens due to lack of deep semantic understanding of an event. Adding more temporal nodes to Nature Design could help to improve the accuracy.

\textbf{Word Sense Ambiguity. } A word is linked to a wrong node due to sense ambiguity. For example, in "you are going to get a light coffee", "light" is incorrectly linked to \texttt{\#emo\_pos}, as "light" here describes the flavor of coffee. Our methods have no ability to determine which sense of a word should be used in an event. A possible cure might be that MEA-DAGs are built for each sense of a word, and then MEA-DAG merging is implemented between sense and context. A proper sense could be merged smoothly into the context. 

\textbf{Wrong Belief. } A word is wrongly linked to emotional or feeling nodes in Nurture Belief. For instance, the words "different" and "raw" are incorrectly connected to \texttt{\#experience\_feeling\_pos}. This happens as SentiWordNet has classification errors, which could not be thoroughly filtered out by LLM. This type of error brings up an interesting research topic: automatic error-correction mechanism. In the course of human development, numerous beliefs are established about this world, some of which are false. Through subsequent experiences, they consistently reinforce correct beliefs and fix wrong beliefs. This mechanism should work perfectly for this kind of error.

\textbf{Wrong Past Action. } Although an action event happens in the past, they are actually driven by how well the food need is met. For instance, "I had to take one star off" is triggered by the dissatisfaction of need. Judging a past action only by tense is not enough. Adding more temporal nodes to Nature Design could help this situation. 

\textbf{Negation Loss. } Events are linked to wrong nodes due to failure of capturing negation. "It just failed to deliver" expresses a negative meaning of action, which is hard to capture unless the semantic meaning of "fail" is incorporated in Nature Design. One solution is adding a layer of nodes which is specifically responsible for dealing with negation and other logic operations.

\subsection{Review Length Effect}
Depending on whether the number of sentences contained in a review is less than 5, the test set is splitted into a short-test set and a long-test set. By comparing the error differences between the short-test and the long-test, we investigate the effect of review length on accuracy. 

Table~\ref{eval_error_anal} shows the comparison result. Incorrectness rate of the short-test set is 29.8\%, while the long-test set has 43.4\%, which indicates that our methods are not well-suited for processing lengthy reviews. Top three error types of the short-test are Wrong Belief, Word Sense Ambiguity and Event Linking Loss, while the long-test are Event Linking Loss, ASER Extraction Loss and Wrong Subsequent Action. From the shift of top three errors, we find that the main bottlenecks of processing long reviews lie in lack of rich nodes and links in Nature Design, as well as lack of comprehensive and in-depth understanding of a sentence.

\section{Conclusion}

We compute MEA-DAGs to understand the relationships among motivations, emotions and actions from natural language texts. Nature Design is novelly introduced to imitate human's nature, and Nurture Belief connects outside world and human's nature. Our methods are white-box and don’t rely on huge annotation resources. Error analysis is implemented to identify the main problems and find possible directions for further optimization.

\bibliography{custom}
\bibliographystyle{plainnat}

\clearpage
\appendix
\section{Error Examples}
\label{sec:appendix_a}
\subsection{Event Linking Loss}

\begin{figure*}[h!]
  \includegraphics[width=2\columnwidth]{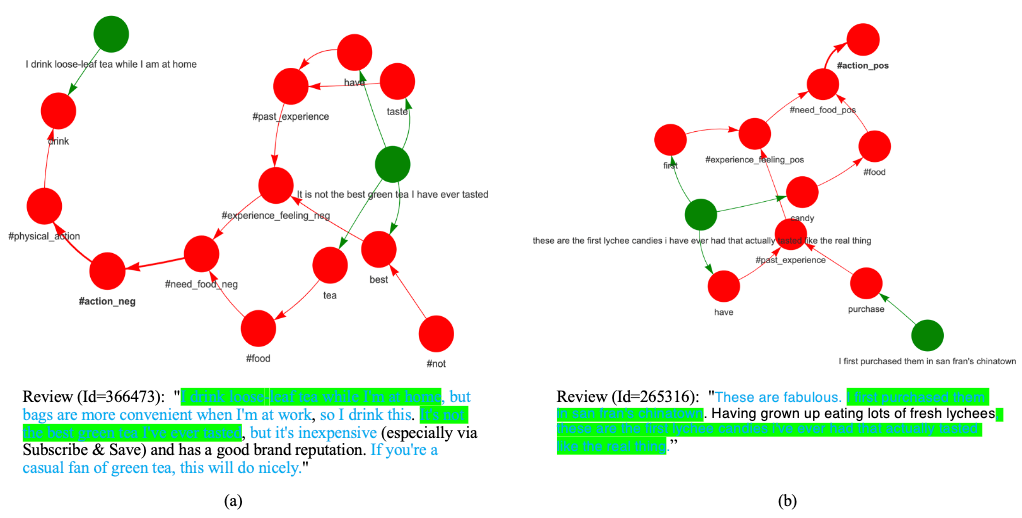}
  \caption{Events extracted by ASER are in blue color. We use a green font background to highlight the events incorporated in MEA-DAG. (a): Critical events "bags are more convenient when I'm at work", "it's inexpensive" are not included in MEA-DAG. (b): Critical event "these are fabulous" are not included in MEA-DAG.  }
  \label{e1_2}
\end{figure*}

Figure~\ref{e1_2} shows two examples whose MEA-DAGs lose critical information of events, resulting that the revealed relationship is not complete in logic sense.

\subsection{ASER Extraction Loss}

\begin{figure*}[h!]
  \includegraphics[width=2\columnwidth]{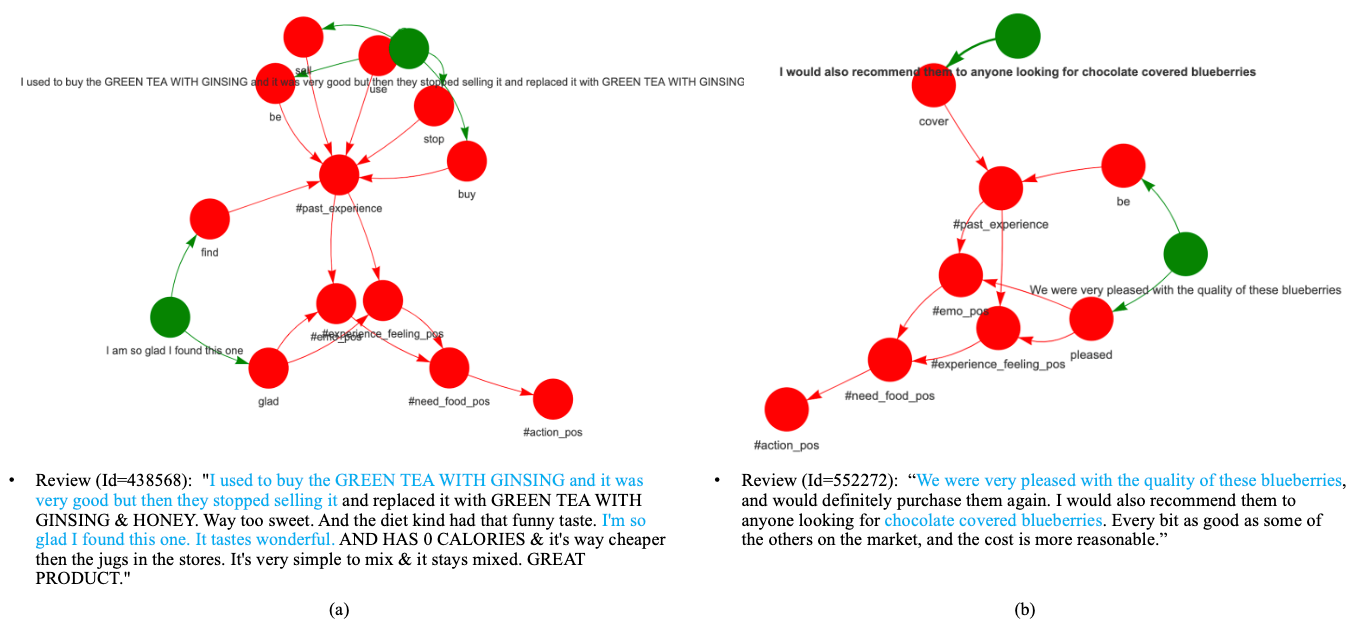}
  \caption{Texts in blue color are the events extracted by ASER. (a): Critical events "the diet kind had that funny taste", "it's way cheaper then the jugs in the stores" and "it's very simple to mix \& it stays mixed" are not captured by ASER. (b): Critical events "would definitely purchase them again", "I would also recommend them" and "the cost is more reasonable" are not captured by ASER.}
  \label{e2_1}
\end{figure*}

Figure~\ref{e2_1} presents two examples in which ASER couldn't extract critical events from review texts. As a result, the generated MEA-DAG is incomplete.

\subsection{Wrong Subsequent Action}
\begin{figure*}[h!]
  \includegraphics[width=2\columnwidth]{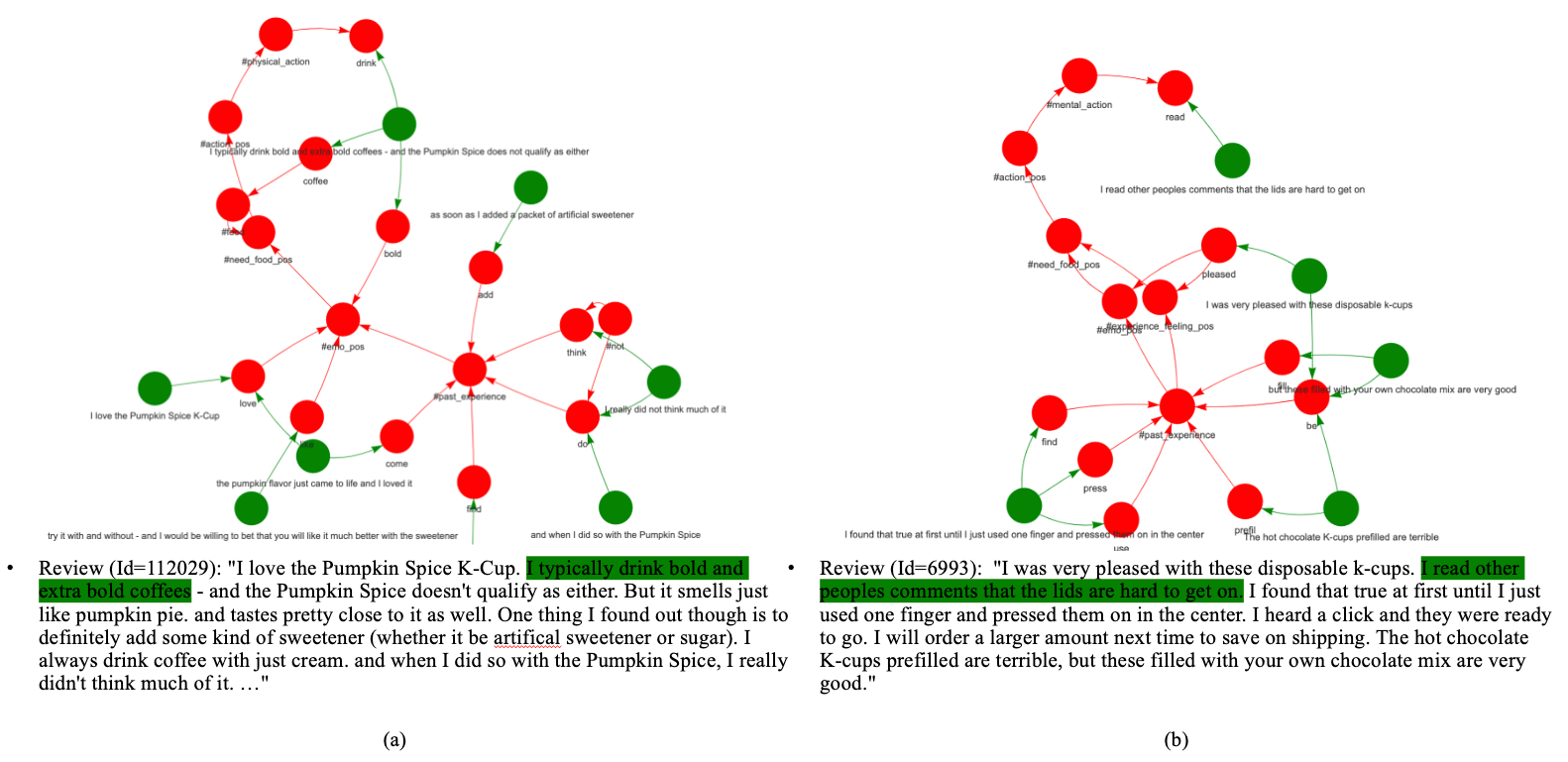}
  \caption{Examples with Wrong Subsequent Action error. We use a green font background to highlight the wrong events. (a): "I typically drink bold extra bold coffee" is linked to \texttt{\#physical\_action}. However, it's not driven by \texttt{\#need\_food\_pos}, as the word "typical" indicates that it's a habitual action. (b): "I read other peoples comments that the lids are hard to get on" is linked to \texttt{\#mental\_action}. However, it should be linked to \texttt{\#past\_experience}. Considering its context, "read" in this event is in the past tense, representing an action that occurred in the past.}
\label{e3_1}
\end{figure*}

Figure~\ref{e3_1} presents two examples in which action events are wrongly linked to the children nodes of \texttt{\#action\_pos}. If the contexts and timeline of events are considered, they should be linked to \texttt{\#past\_experience}.

\subsection{Word Sense Ambiguity}
\begin{figure*}[h!]
  \includegraphics[width=2\columnwidth]{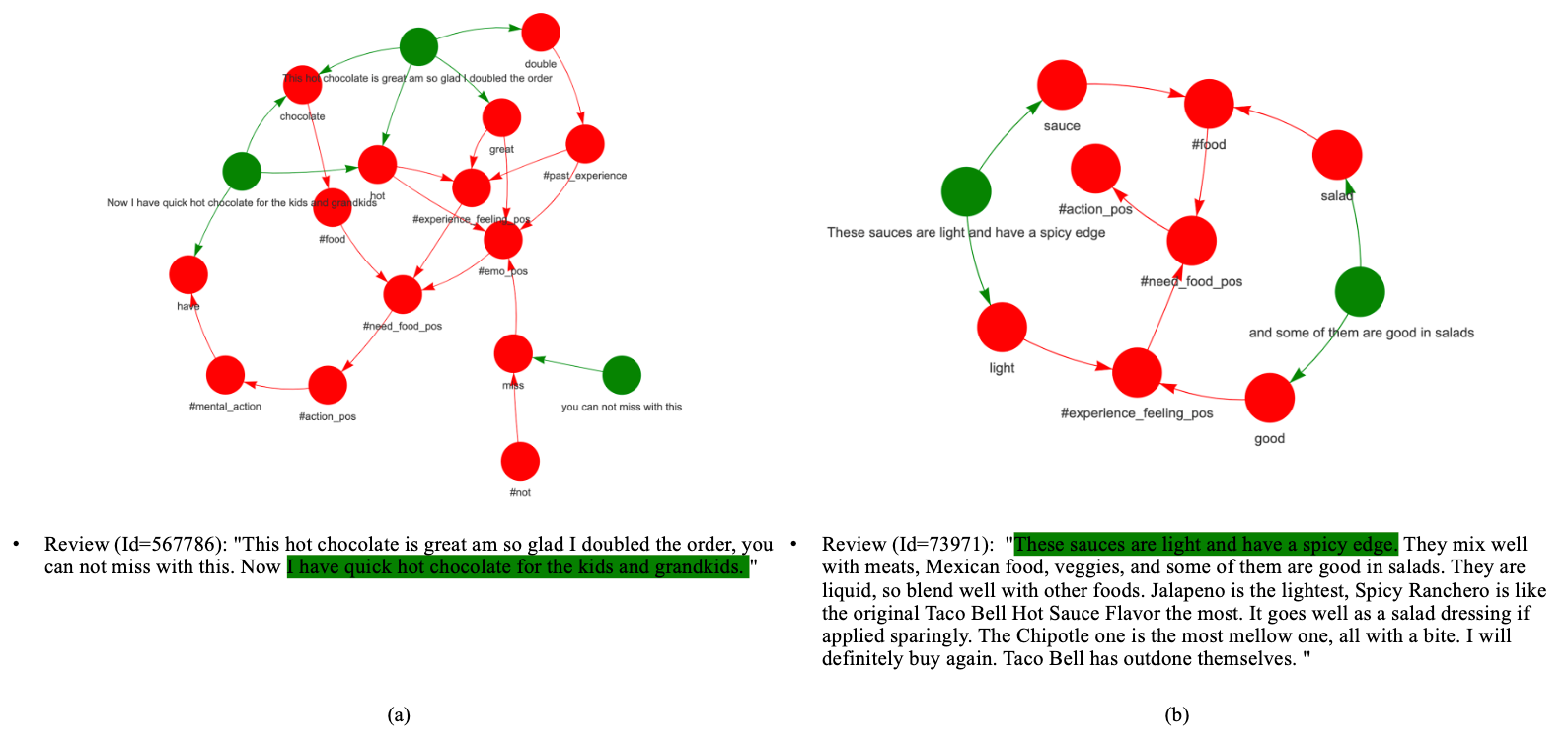}
  \caption{Examples with Word Sense Ambiguity error. We use a green font background to highlight the wrong events. (a): In the event "I have quick hot chocolate for the kids and grandkids", "hot" is an objective description of food, not bearing an positive emotion or feeling sense.  (b): In the event "these sauces are light and have a spicy edge", "light" is  wrongly linked to \texttt{\#experience\_feeling\_pos}, as it describes sauce taste, not a feeling.}
\label{e4_1}
\end{figure*}

Figure~\ref{e4_1} presents two examples in which words are wrongly linked to feeling or emotion nodes, as our methods have no ability to determine the sense of a word given its context. 

\subsection{Wrong Belief}
\begin{figure*}[h!]
  \includegraphics[width=2\columnwidth]{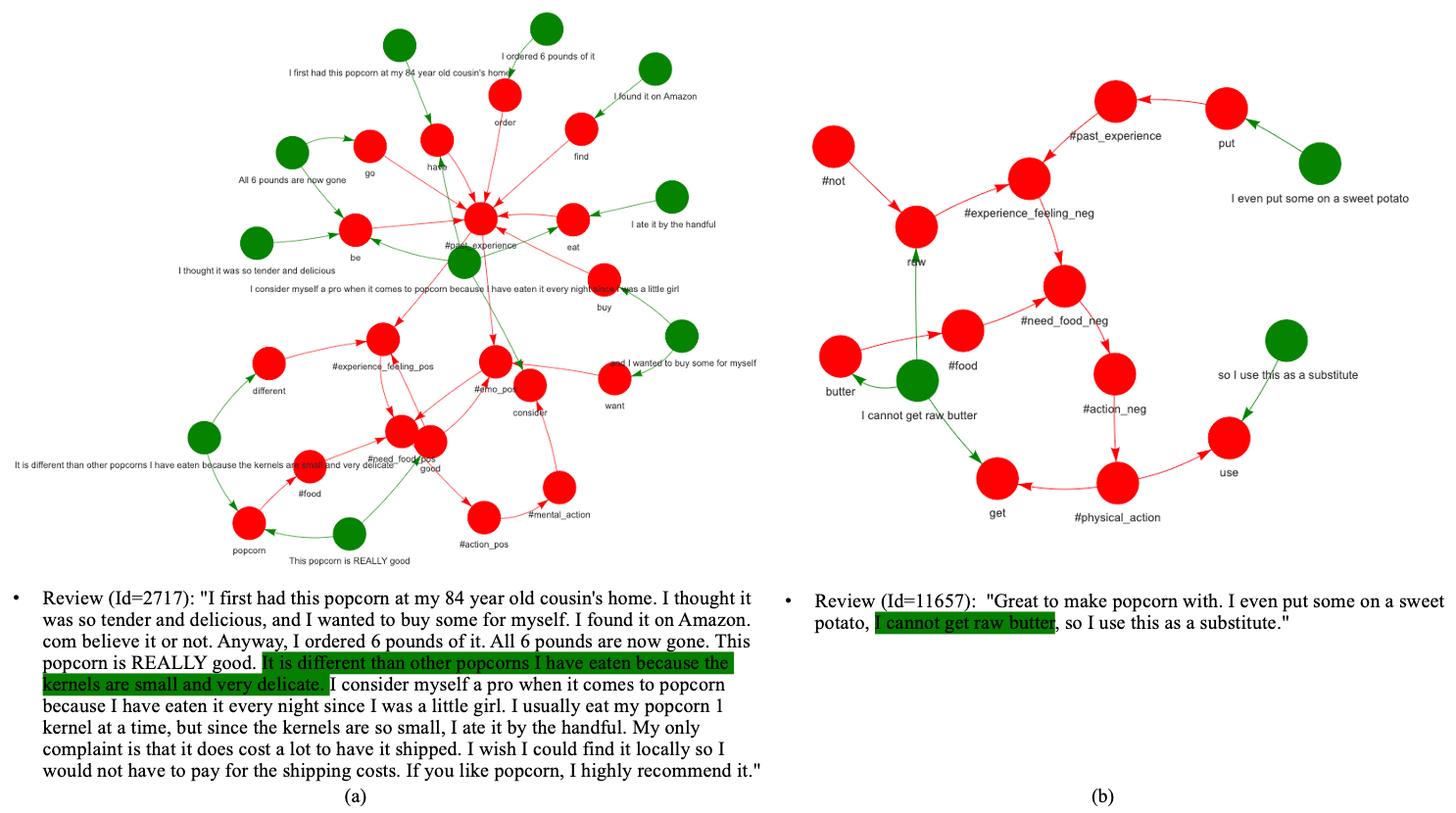}
  \caption{Examples with Wrong Belief error. We use a green font background to highlight the wrong events. (a): The word "different" is incorrectly linked to \texttt{\#experience\_feeling\_pos}. (b): The word "raw" is incorrectly linked to \texttt{\#experience\_feeling\_pos}.}
\label{e7_1}
\end{figure*}

Figure~\ref{e7_1} presents two examples in which events are linked to incorrect nodes due to errors in Nurture Belief.

\subsection{Wrong Past Action}

\begin{figure*}[h!]
  \includegraphics[width=2\columnwidth]{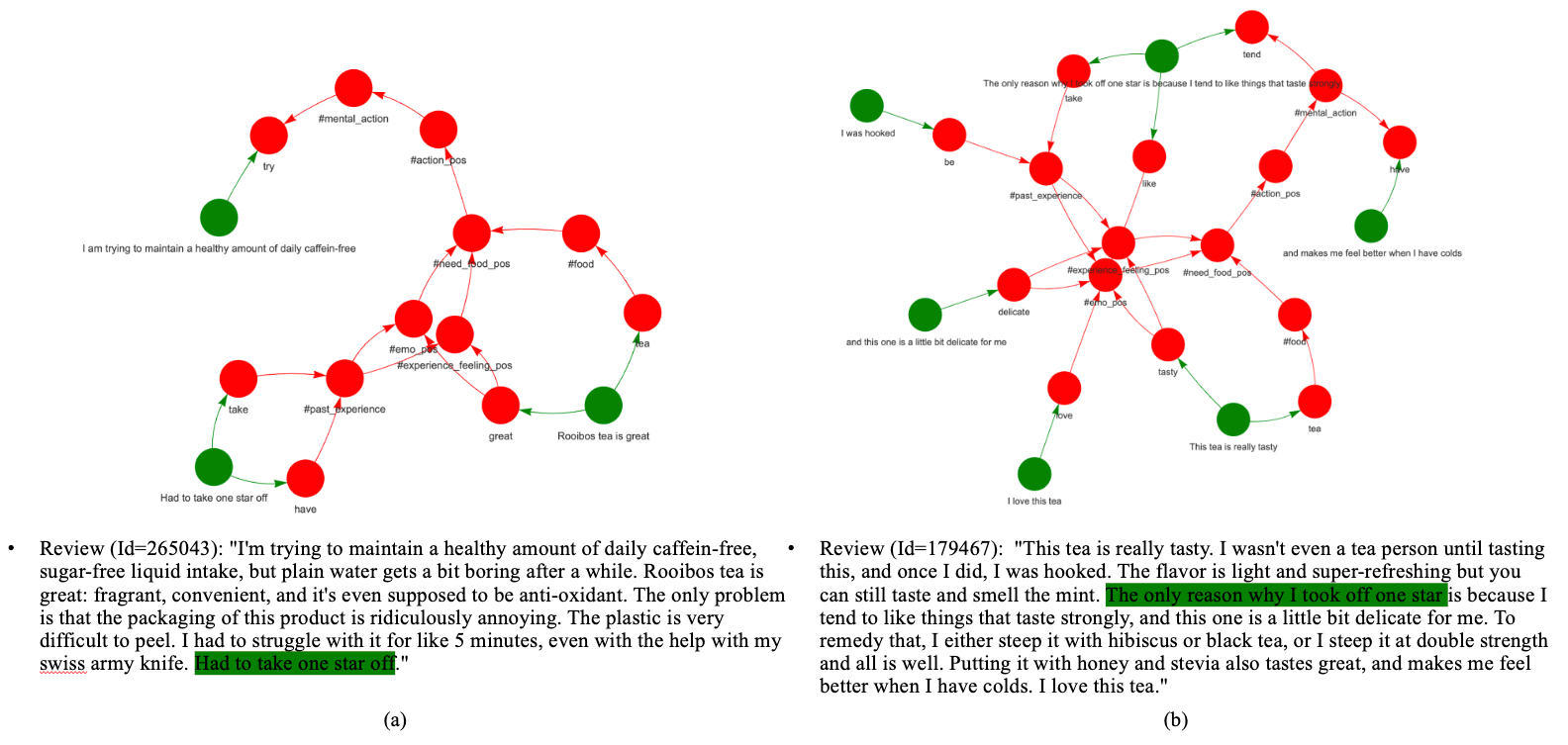}
  \caption{Examples with Wrong Past Action error. We use a green font background to highlight the wrong events. (a): The event "Had to take one star off" happened due to dissatisfaction with food. Therefore, it should be linked to \texttt{\#physical\_action}. (b): The event "The only reason why I took off one star " is driven by dissatisfaction with food, not a factor that affects whether the need is met.}
\label{e5_1}
\end{figure*}

Figure~\ref{e5_1} presents two examples in which events are wrongly linked to \texttt{\#past\_experience}, as they are not factors that affect whether the need is met. In fact, they are the result of food need not being met.

\subsection{Negation Loss} 
\begin{figure*}[h!]
  \includegraphics[width=2\columnwidth]{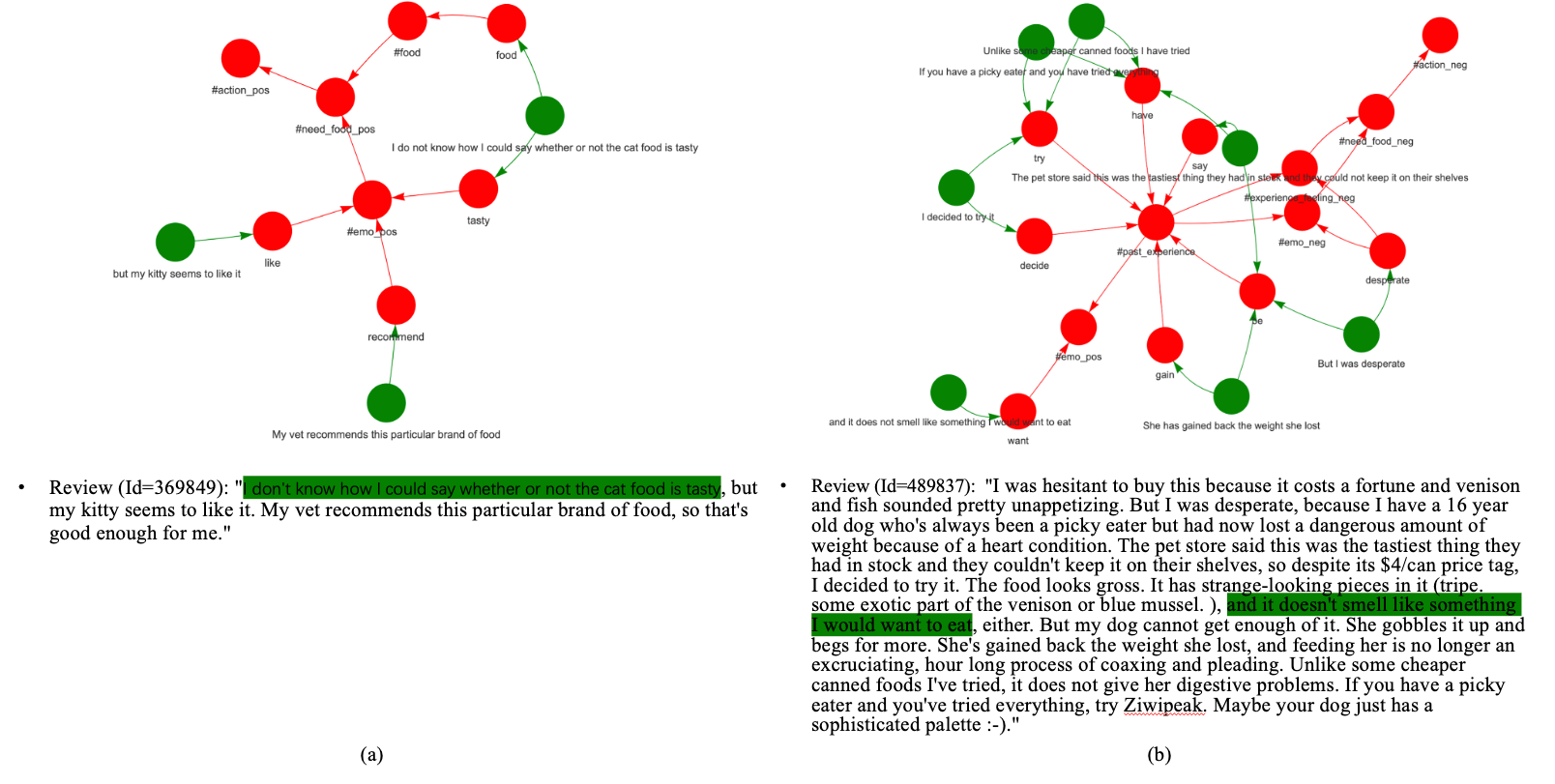}
  \caption{Examples with Negation Loss error. We use a green font background to highlight the wrong events. (a): The event "I don't know how I could say whether or not the cat food is tasty" is wrongly linked to \texttt{\#emo\_pos} due to failure of capturing "not". Although "tasty" has a sense of positive emotion, "not" changes its meaning to opposite side. (b): The event "it doesn't smell like something I would want to eat" is wrongly linked to \texttt{\#emo\_pos}, as "doesn't" is not captured in the MEA-DAG.}
\label{e6_1}
\end{figure*}

Figure~\ref{e6_1} presents two examples in which events are wrongly linked to \texttt{\#emo\_pos}, as the negation expressions "don't", "whether or not" and "doesn't" are not captured by our methods. 

\end{document}